\documentclass[11pt,a4paper]{article}
\usepackage[hyperref]{acl}
\usepackage{times}
\usepackage{latexsym}
\usepackage{booktabs}
\usepackage{dirtytalk}
\usepackage{graphicx}
\usepackage{listings,multicol}
\usepackage{courier}
\usepackage{tikz}
\usepackage{stfloats}
\usepackage{multirow}

\usepackage{arabtex}
\renewenvironment{abstract}%
         {\centerline{\large\bf Abstract}%
          \begin{list}{}%
             {\setlength{\rightmargin}{0.6cm}%
              \setlength{\leftmargin}{0.6cm}}%
           \item[]\ignorespaces}%
         {\unskip\end{list}}

\usepackage{cjhebrew} % http://mirrors.ibiblio.org/CTAN/language/hebrew/cjhebrew/cjhebrew.pdf
\usepackage{array}
\newcolumntype{H}{>{\setbox0=\hbox\bgroup}c<{\egroup}@{}}

\graphicspath{ {.} }

\usepackage{microtype}

\newcommand\sysname{\textsc{Nakdimon}}

\newcommand{\iflong}[1]{}

\title{Restoring Hebrew Diacritics Without a Dictionary}

\author{Elazar Gershuni \\
  \phantom{1} \\
  Technion -- Israel Institute of Technology \\
  Haifa, Israel \\
  \texttt{elazarg@gmail.com} \\\And
  Yuval Pinter \\
  Department of Computer Science \\
  Ben-Gurion University of the Negev \\
  Beer Sheva, Israel \\
  \texttt{uvp@cs.bgu.ac.il} \\}

\date{}

\begin{document}
\maketitle

\begin{abstract}
We demonstrate that it is feasible to accurately diacritize Hebrew script without any human-curated resources other than plain diacritized text.

We present \sysname{}, a two-layer character-level LSTM, that performs on par with much more complicated curation-dependent systems, across a diverse array of modern Hebrew sources.
The model is accompanied by a training set and a test set, collected from diverse sources.
\end{abstract}

% https://openreview.net/forum?id=gXOlHDiNcs2&noteId=De5NV0zSdS

\section{Introduction}
\label{sec:intro}

The vast majority of modern Hebrew texts are written in a letter-only version of the Hebrew script, one which omits the diacritics present in the full diacritized, or \emph{dotted} variant.\footnote{Also known as \emph{pointed} text, or via the Hebrew term for the diacritic marks, \emph{nikkud/niqqud}.}
Since most vowels are encoded via diacritics, the pronunciation of words in the text is left underspecified, and a considerable mass of tokens becomes ambiguous.
This ambiguity forces readers and learners to infer the intended reading using syntactic and semantic context, as well as common sense~\cite{bentin1987processing,abu2001role}.
In NLP systems, recovering such signals is difficult, and indeed their performance on Hebrew tasks is adversely affected by the presence of undotted text~\cite{shacham-wintner-2007-morphological,goldberg-elhadad-2010-easy,tsarfaty-etal-2019-whats}.

As an example, the sentence in~\autoref{tab:ex} (a) will be resolved by a typical reader as (b) in most reasonable contexts, knowing that the word \say{softly} may characterize landings.
In contrast, an automatic system processing Hebrew text may not be as sensitive to this kind of grammatical knowledge and instead interpret the undotted token as the more frequent word in (c), harming downstream performance.

\begin{table}
    % \small
    \centering
    \begin{tabular}{lc} \toprule
        % \cjRL{d*:bAriyM +sErA.siytiy lOmAr} \\
        % \cjRL{dbryM /sr.syty lwmr} \\ \bottomrule
        \multirow{3}{*}{(a)} & \cjRL{hm.tws nxt brkwt} \\
        & \sffamily{\small hamatos naxat ????} \\
        & `The plane landed (unspecified)' \\
        \multirow{3}{*}{(b)} & \cjRL{ham*A.tOs nAxat b*:rak*Ut}  \\
        & \sffamily{\small hamatos naxat b-rakut} \\
        & `The plane landed softly' \\
        \multirow{3}{*}{(c)} & \cjRL{ham*A.tOs nAxat b*:rAkOt}  \\
        & \sffamily{\small hamatos naxat braxot} \\
        & `The plane landed congratulations' \\ \bottomrule
    \end{tabular}
    \caption{An example of an undotted Hebrew text (a) (written right to left) which can be interpreted in at least two different ways (b,c), dotted and pronounced differently, but only (b) makes grammatical sense.
    % Disambiguation is impossible without long-range context or extralinguistic knowledge.
    }% \todo{Figure out font issues: \url{http://texdoc.net/texmf-dist/doc/latex/babel/babel.pdf}; or just insert a .png.}}
    \label{tab:ex}
\end{table}

One possible way to overcome this problem is by adding diacritics to undotted text, or \emph{dotting}, implemented using data-driven algorithms trained on dotted text.
Obtaining such data is not trivial, even given correct pronunciation:
the standard Tiberian diacritic system contains several sets of identically-vocalized forms,
so while most Hebrew speakers easily read dotted text, they are unable to produce it.
Moreover, the process of manually adding diacritics in either handwritten script or through digital input devices is mechanically cumbersome.
Thus, the overwhelming majority of modern Hebrew text is undotted, and manually dotting it requires expertise.
The resulting scarcity of available dotted text in modern Hebrew contrasts with Biblical and Rabbinical texts which, while dotted, manifest a very different language register.
This state of affairs allows individuals and companies to offer dotting as paid services, either by experts or automatically, e.g.~the Morfix engine by Melingo.\footnote{\url{https://nakdan.morfix.co.il/}}
Such usage practices also force a disconnect in the NLP pipeline, requiring an API call into an external service whose parameters cannot be updated.

Existing computational approaches to dotting are manifested as complex, multi-resourced systems which perform morphological analysis on the undotted text and look undotted words up in hand-crafted dictionaries as part of the dotting process.
Dicta's Nakdan~\cite{shmidman-etal-2020-nakdan}, the current state-of-the-art, applies such methods in addition to applying multiple neural networks over different levels of the text, requiring manual annotation not only for dotting but also for morphology.
Among the resources it uses are a diacritized corpus of ~3M tokens and a POS-tagged corpus of ~300K tokens. 
Training the model takes several weeks.\footnote{Private communication.}

In this work, we set out to simplify the dotting task as much as possible to standard modules.
We introduce a large corpus of semi-automatically dotted Hebrew, collected from various sources, and use it to train an RNN-based model.
Our system, \sysname{}, accepts the undotted character sequence as its input, consults no external resources or lexical components, and produces diacritics for each character, resulting in dotted text whose quality is comparable to that of the commercial Morfix, on both character-level and word-level accuracy.
Our model is easy to integrate within larger systems that perform end-to-end Hebrew processing tasks, as opposed to the existing proprietary dotters. %, which require API calls.
To our knowledge, this is the first attempt at a \say{light} model for Hebrew dotting since early HMM-based systems~\cite{kontorovichproblems,gal-2002-hmm}.
\iflong{In addition, we evaluate the utility of using dotted text in downstream tasks on Hebrew.
This utility has not been achieved until now, and in cases of deriving parallelisms between Hebrew and Arabic text even shown to e detrimental~\cite{chew-abdelali-2008-effects}.
We show that\dots}

We introduce a novel test set for Modern Hebrew dotting, derived from larger and more diverse sources than existing datasets.
In experiments over our dataset, we show that our system is particularly useful in the main use case of modern dotting, which is to convey the desired pronunciation to a reader, and that the errors it makes should be more easily detectable by non-professionals than Dicta's.\footnote{The system is available at 
% \url{anonymized.org}.}
\url{https://nakdimon.org}, and the source code is a available at \url{https://github.com/elazarg/nakdimon}.}
%

% \section{Hebrew Diacritization}
% \label{sec:theory}

% \input{arch_graph}

\section{Task and Datasets}
\label{sec:dataset}

\subsection{Dotting as Sequence Labeling}
The input to the dotting task consists of a sequence of characters. Each of the characters is assigned three values, from three separate diacritic categories: one category for the dot distinguishing \textit{shin} (\cjRL{+s}) from \textit{sin} (\cjRL{,s}), two consonants sharing a base character \cjRL{/s}; another for the presence of \emph{dagesh}/\emph{mappiq}, a central dot affecting pronunciation of some consonants, e.g.~\cjRL{p*} /p/ from \cjRL{p|} /f/, but also present elsewhere; and one for all other diacritic marks, which mostly determine vocalization, e.g. \cjRL{dA} /da/ vs. \cjRL{dE} /de/.
Diacritics of different categories may co-occur on single letters, e.g. \cjRL{+s*A}, or may be absent altogether.

\paragraph{Full script}
Hebrew script written without intention of dotting typically employs a compensatory variant known colloquially as full script (\emph{ktiv male}, \cjRL{ktyb ml'}), which adds instances of the letters \cjRL{y} and \cjRL{w} in some places where they can aid pronunciation, but are incompatible with the rules for dotted script.
In our formulation of dotting as a sequence tagging problem, and in collecting our test set from raw text, these added letters may conflict with the dotting standard.
For the sake of input integrity, and unlike some other systems, we opt not to remove these characters, but instead employ a dotting policy consistent with full script.
See Appendix~\ref{app:male} for further details.

\subsection{Training corpora}
\label{sec:train}
Dotted modern Hebrew text is scarce, since speakers usually read and write undotted text, with the occasional diacritic added for disambiguation when context does not suffice.
As we are unaware of legally-obtainable dotted modern corpora, we use a combination of dotted pre-modern texts as well as automatically and semi-automatically dotted modern sources to train \sysname{}:

The \textsc{pre-modern} portion is obtained from two main sources: A combination of late pre-modern text from Project Ben-Yehuda, mostly texts from the late 19th century and the early 20th century;\footnote{ \mbox{\url{https://benyehuda.org}}} rabbinical texts from the medieval period, the most important of which is Mishneh Torah (obtained from Project Mamre);\footnote{\mbox{\url{https://mechon-mamre.org}}} and 23 short stories from the short story project.\footnote{\mbox{\url{https://shortstoryproject.com/he/}}}
This portion contains roughly 1.81M Hebrew tokens, most of which are dotted, with a varying level of accuracy, varying dotting styles, and varying degree of similarity to Modern Hebrew.

The \textsc{automatic} portion contains 547 short stories taken from the short story project.
The stories are dotted using Dicta \emph{without} manual validation.
The corpus contains roughly 1.27M Hebrew tokens.

Lastly, the \textsc{modern} portion contains manually collected text in Modern Hebrew, mostly from undotted sources, which we dot using Dicta and follow up by manually fixing errors, either using Dicta's API or via automated scripts which catch common mistakes.
We made an effort to collect a diverse set of sources: news, opinion columns, paragraphs from books, short stories, Wikipedia articles, governmental publications, blog posts and forums expressing various domains and voices, and more.
% As our dotting guidelines aim for readability and faithfulness to modern writing conventions, we were able to create a corpus large enough with limited expertise in dotting.
Our \textsc{modern} corpus contains roughly 326K Hebrew tokens, and is much more consistent and similar to the expectation of a native Hebrew speaker than the \textsc{pre-modern} or the \textsc{automatic} corpora, and more accurately dotted than the \textsc{automatic} corpus.
The sources and statistics of this dataset are presented in \autoref{tab:data}.

\begin{table}
    \centering
    \small
    \begin{tabular}{cllrr}
    \toprule
            & Genre      & Sources        & \# Docs & \# Tokens \\
\midrule
            & Wiki       & Dicta test set &     22 &      5,862 \\
            & News       & Yanshuf        &     78 &     11,323 \\
{$\dagger$} & Literary   & Books, forums  &    129 &     73,770 \\
{*}         & Official   & gov.il         &     24 &     20,181 \\
{*}         & News / Mag & Online outlets &    137 &     92,151 \\
{*}         & User-gen.  & Blogs, forums  &     63 &     60,673 \\
{*}         & Wiki       & he.wikipedia   &     40 &     62,723 \\
\midrule
            & Total      &                &    493 &    326,683 \\
        \bottomrule
    \end{tabular}
    \caption{Data sources for our \textsc{modern} Hebrew training set. Rows marked with * were automatically dotted via the Dicta API and corrected manually.
    Rows with $\dagger$ were dotted at low quality, requiring manual correction.
    The rest were available with professional dotting.}
    \label{tab:data}
\end{table}

\subsection{New test set}
\label{sec:test}
\newcite{shmidman-etal-2020-nakdan} provide a benchmark dataset for dotting modern Hebrew documents.
However, it is relatively small and non-diverse: all 22 documents in the dataset originate in a single source, namely Hebrew Wikipedia articles. % 5,862 tokens

Therefore, we created a new test set\footnote{\url{https://github.com/elazarg/hebrew_diacritized/tree/master/test_modern}} from a larger variety of texts, including high-quality Wikipedia articles and edited news stories, as well as user-generated blog posts.
This set consists of ten documents from each of eleven sources (5x Dicta's test set), and totals 20,474 Hebrew tokens, roughly 3.5x Dicta's. We use the same technique and style for dotting this corpus as we do for the \textsc{modern} corpus (\S\ref{sec:train}), but the documents were collected in different ways.

% Both train and test data are publicly available at \url{github.com/elazarg/hebrew_diacritized}.

% \input{arch_graph}

\section{Nakdimon}
\label{sec:model}
\sysname{} embeds the input characters and passes them through a two-layer Bi-LSTM~\cite{hochreiter1997long}.
The LSTM output is fed into a single linear layer, which then feeds three linear layers, one for each diacritic category (see \S\ref{sec:dataset}).
Each character then receives a prediction for each category independently and all predicted marks are added to it as output.
% \autoref{fig:arch} shows the overall architecture of the system.

Decoding is performed greedily, with no validation of readability or any other dependence between character-level decisions.

The input is pre-processed by removing all but Hebrew characters, spaces and punctuation; digits are converted to a dedicated symbol, as are Latin characters. All existing diacritic marks are stripped, and each document is split into chunks % of at most 80 characters
bounded at whitespace, ignoring sentence boundaries.

We train \sysname{} first over \textsc{pre-modern}, then over the 
\textsc{automatic} corpus, and then by over the \textsc{modern} corpus. During training, the loss is the sum of the cross-entropy loss from all three categories.
Trivial decisions, such as the label for the \textit{shin}/\textit{sin} diacritic for any non-\cjRL{/s} letter, are masked.

Tuning experiments are detailed in Appendix~\ref{app:dev}; an evaluation of a preliminary version of \sysname{} over the Dicta test set is in Appendix~\ref{app:results}, and Hyperparameters are detailed in Appendix~\ref{app:hyper}.

\section{Experiments}
\label{sec:exp}

% Following \newcite{shmidman-etal-2020-nakdan}, w
We compare the performance of \sysname{} on our new test set (\S\ref{sec:test}) against Dicta,\footnote{Version 4.0, wordlist version 43.} Snopi,\footnote{\url{http://nakdan.com/Nakdan.aspx}} and Morfix~\cite{kamir-etal-2002-comprehensive}.
% In order to report compatible findings with those in previous work, we present results on both the Dicta test, adapted to full script, and on our new test set.
as well as a \textsc{Majority} baseline which returns the most common dotting for each word seen in our full training set.

\paragraph{Metrics} We report four metrics:
\textbf{decision accuracy (\textsc{dec})} is computed over the entire set of individual possible decisions: \emph{dagesh}/\emph{mappiq} for letters that allow it, \textit{sin}/\textit{shin} dot for the letter \cjRL{/s}, and all other diacritics for letters that allow them;
\textbf{character accuracy (\textsc{cha})} is the portion of characters in the text that end up in their intended final form (which may combine two or three decisions, e.g. \emph{dagesh} + vowel);
\textbf{word accuracy (\textsc{wor})} is the portion of words with no mistakes; and
\textbf{vocalization accuracy (\textsc{voc})} is the portion of words where any dotting errors do not cause incorrect pronunciation among mainstream Israeli Hebrew speakers.\footnote{These are: the \textit{sin}/\textit{shin} dot, vowel distinctions across the a/e/i/o/u/null sets, and \emph{dagesh} in the \cjRL{b}/\cjRL{k|}/\cjRL{p|} characters. We do not distinguish between \emph{kamatz gadol}/\emph{kamatz katan}, and \emph{schwa} is assumed to always be null.}

\begin{table}
    \centering
    \small
    \begin{tabular}{lcccc}
        \toprule
        System & \textsc{dec} & \textsc{cha} & \textsc{wor} & \textsc{voc}\\
        \midrule
        \textsc{Majority} & 93.79 & 90.01 & 84.87 & 86.19\\
        \midrule
        \textsc{Snopi} & 91.29 & 85.84 & 76.45 & 78.91 \\
        \textsc{Morfix} & 96.84 & 94.92 & 90.38 & 92.39 \\
        \textsc{Dicta}  & \textbf{97.95} & \textbf{96.77} & \textbf{94.11} & \textbf{94.92} \\
        \midrule
        \sysname{} & 97.91 & 96.37 & 89.75 & 91.64 \\
        \bottomrule
    \end{tabular}
    \caption{Document-level macro \% accuracy.}
    \label{tab:res_short}
\end{table}

\subsection{Results}
We provide document-level macro-averaged accuracy percentage results for a single run over our test set in \autoref{tab:res_short}.
All systems, except Snopi, substantially outperform the majority-dotting baseline on all metrics.
\sysname{} outperforms Morfix on character-level metrics but not on word-level metrics, mostly since Morfix ignores certain words altogether, incurring errors on multiple characters.

% This is less impressive now:
We note the substantial improvement our model achieves on the \textsc{voc} metric compared to the \textsc{wor} metric: 18.43\% of word-level errors are attributable to vocalization-agnostic dotting, compared to 13.80\% for Dicta and 10.41\% for Snopi (but 20.91\% for Morfix).
Considering that the central use case for dotting modern Hebrew text is to facilitate pronunciation to learners and for reading, and that undotted homograph ambiguity typically comes with pronunciation differences, we believe this measure to
be no less important than \textsc{wor}.

Results on Dicta's test set~\cite{shmidman-etal-2020-nakdan} are presented in Appendix~\ref{app:results}.

\subsection{Error analysis}
In \autoref{tab:errors} we present examples of words dotted incorrectly, or correctly, only by \sysname{}, compared with Morfix and Dicta.
The largest category for \sysname{}-only errors ($\sim$18\% of 90 sampled) are ones where a fused preposition+determiner character is dotted to only include the preposition, perhaps due to its inability to detect the explicit determiner clitic \cjRL{h} in neighboring words, on which the complex systems apply morphological segmentation.
In other cases ($\sim$15\%), \sysname{} creates unreadable vocalization sequences, as it has no lexical component and is decoded greedily.
These types of errors are more friendly to the typical use cases of a dotting system, as they are likely to stand out to a reader.
% Other recurring error types include named entities and errors at sentence boundaries, which likely stem from lack of context.
In contrast, a large portion of cases where only \sysname{} was correct ($\sim$13\% of 152) are foreign names and terms.
This may be the result of such words not yet appearing in dictionaries, or not being easily separable from an adjoining clitic, while character-level information can capture pronunciation patterns from similar words (e.g. \cjRL{.tElEpwon} `telephone', for the example \cjRL{h'yypwn}). % Of all 70 cases we tagged as \say{foreign}, Morfix got only 7 right and \sysname{} 26 (Dicta 40).
% Dicta's errors mostly constitute selection of the wrong \emph{in-vocabulary} word in a context, or a wrong inflection of a verb.

\begin{table}
    \small
    \centering
    \begin{tabular}{ccc} \toprule
        Context & Correct & Incorrect \\
        \midrule
        .\cjRL{w.sryk lhstkl lh b`ynyym} \dots & \cjRL{b*A`eynayiym} & \cjRL{b*:`eynayiym} \\
        \multicolumn{3}{c}{`\dots{}and we need to look her \textbf{in the eyes (/in eyes)}.'}
        \vspace{4pt}  \\
        \dots \cjRL{y`nw lK bsblnwt} \dots & \cjRL{lAK:} & \cjRL{l:K:} \\
        \multicolumn{3}{c}{`\dots\textbf{you.sg.f (/unreadable)} will be answered patiently\dots'} \\
        \midrule
        \dots \cjRL{m/stm/sy h'yypwn hr'/swnym} \dots & \cjRL{hA'ay:ypwon} & \cjRL{hA'iyyp*won} \\
        \multicolumn{3}{c}{`\dots{}the first \textbf{iPhone (/\textit{ee-pon})} users\dots'} \\
        \bottomrule
    \end{tabular}
    \caption{Examples of words dotted incorrectly (top) or correctly (bottom) only by \sysname{}.}
    \label{tab:errors}
\end{table}

\paragraph{OOVs} To further quantify the strengths of \sysname{}'s architecture and training abilities, we evaluate the systems' results pertaining only to those words in the test set which do not appear in our training sets.
We follow common practice by calling them OOVs (\say{out of vocabulary}), but emphasize that \sysname{} does not consult an explicit vocabulary, and the other systems are not evaluated against their own vocabularies (which are unknown to us).

We find that \sysname{}'s performance on this subset is substantially \textbf{worse} compared with the other systems than on the full set: 15 percentage points below Dicta and seven below Morfix on the \textsc{voc} metric (see full results in \autoref{app:results}).

These results might be counter-intuitive considering the proven utility of character-level models in OOV contexts~\cite[e.g.,][]{plank-etal-2016-multilingual}, and so we offer several possible explanations:
First, many \say{OOVs} consist in fact of known words coupled with an unseen combination of prefix clitics and/or suffix possessive markers, which other systems explicitly remove using morphological analyzers before dotting.
Second, mirroring the last finding from the overall analysis, some \say{OOVs} are proper names which appear in dictionaries but are absent from the training set, due to corpus effects such as time and domain, or simply chance.

\iflong{
    \section{Parsing Dotted Text}
    (TODO: run POS tagging using pre-trained embeddings, lookup via undotted)
    [POS results, 40 epochs, no embeddings: dotted 0.9260 undotted 0.9325 mimick paper 0.9659]
    \label{sec:parse}
    (TODO: Morphological reinflection~\cite{aharoni-goldberg-2017-morphological})
}

\section{Related Work}
\label{sec:related}
Existing work on diacritizing Hebrew is not common, and all efforts build on word-level features.

\newcite{kontorovichproblems} trains an HMM on a vocalized and morphologically-tagged portion of the Hebrew Bible containing 30,743 words, and evaluates the result on a test set containing 2,852 words, achieving $81\%$ WOR accuracy.
Note that Biblical Hebrew is very different from Modern Hebrew in both vocabulary, grammatical structure, and diacritization, and also has many words with unique diacritization.
In our system, we exclude the Bible altogether from the training set, as its inclusion actively hurts performance on the validation set, which consists of Modern Hebrew.

\newcite{tomer2012automatic} designs a diacritization system for Hebrew verbs consisting of a combination of a verb inflection system, a syllable boundary detector, and an SVM model for classifying verb inflection paradigms.
The focus on verbs in a type-level setup makes this work incomparable to ours or to others in this survey.

% morphologically the most complex part of the language. He generates 250,000 inflected, diacritized verbs from a list of 4,000 verbs. A target word is first split into syllables using a rule-based approach, and then the form is looked-up in the list of inflected verbs. Verbs that are not on this list are diacritized using an SVM based on the words and some rule-based corpus-level features.

In Arabic, diacritization serves a comparable purpose to that in Hebrew, but not exclusively: most diacritic marks differentiate consonantal phonemes from each other, e.g.~
{\RL{b}} /b/ vs. 
{\RL{t}} /t/ (which only the sin/shin dot does in Hebrew),
whereas vocalization marks are in a one-to-one relationship with their phonetic realizations, e.g.~only the \textit{fatha} as in
{\RL{ba}} /ba/
encodes the /a/ vowel.

Dictionary-less Arabic diacritization has been attempted using a 3-layer Bi-LSTM~\cite{belinkov-glass-2015-arabic}. % with a sliding window of size 5.
\newcite{abdanah-arabic} use a Bi-LSTM where characters are assigned either one or more diacritic symbols.
Our system differs from theirs by virtue of separating the diacritization categories.
\newcite{mubarak-etal-2019-highly} tackled Arabic diacritization as a sequence-to-sequence problem, tasking the model with reproducing the characters as well as the marks.

\newcite{zalmout-habash-2017-dont} have made the case against RNN-only systems, arguing for the importance of morphological analyzers in Arabic NLP systems.
We concede that well-curated systems may perform better than uncurated ones, particularly on low-resource languages such as Hebrew, but we note that they are difficult to train for individual use cases and are burdensome to incorporate within larger systems.

Diacritics restoration in Latin-based scripts, applicable mostly to European languages, forms a substantially different problem from the one in Hebrew given the highly lexicalized nature of diacritic usage in these languages and the very low rate of characters requiring diacritics.
The state-of-the-art systems in such languages employ transformer models in a sequence-to-sequence setup~\cite{naplava2021diacritics,Stankevicius2022}, supplanting character-RNN sequence prediction architectures reminiscent of ours~\cite{naplava-etal-2018-diacritics}.
Indeed, the authors of this latter work note the only non-European in their dataset, Vietnamese, as a special outlier.

\section{Conclusion}
\label{sec:conc}
Learning directly from plain diacritized text can go a long way, even with relatively limited resources. \sysname{} demonstrates that a simple architecture for diacritizing Hebrew text as a sequence tagging problem can achieve performance on par with much more complex systems.
We also introduce and release a corpus of dotted Hebrew text, as well as a source-balanced test set.

In the future, we wish to evaluate the utility of dotting as a feature for downstream tasks such as question answering, machine translation, and speech generation, taking advantage of the fact that our simplified model can be easily integrated in an end-to-end Hebrew processing system.

\section*{Ethical Considerations}
% \url{https://2021.naacl.org/ethics/faq/#should-i-have-an-ethical-considerations-section-in-my-paper}
We collected the data for our training set and test sets from open online sources, while making sure their terms allow research application and privacy is not impugned.
\sysname{}'s architecture does not encourage memorization of training data and the system is not trained for generating text.

We consider a main use case for our system to be assisting Hebrew learners in reading.
We therefore expect \sysname{} to facilitate life in Israel for immigrants still struggling with Hebrew, among other underprivileged groups.
Automatic dotting can increase inclusion in Hebrew-prominent societies for literacy-challenged individuals, and derivative improvements in text-to-speech applications can assist those with impaired vision.
Lastly, dotting can help researchers with limited understanding of Hebrew access resources in the language.

Hebrew is a gendered language.
Orthographically, in many cases the lack of dots masks gender ambiguity, allowing both masculine and feminine readings for a given word (e.g.~\cjRL{SAlax:t*:} / \cjRL{SAlax:t*A} `you.fem sent' / `you.masc sent').
While well-performing automatic dotting can help alleviate these ambiguities and reduce the amount of potentially prejudiced readings, we recognize the large body of work on gender bias in NLP~\cite{blodgett-etal-2020-language}, including in Hebrew NLP~\cite{moryossef-etal-2019-filling}, and the findings that an imbalanced training set may result in an even more skewed distribution of gender bias in applications~\cite{zhao-etal-2017-men}.
We believe our unlexicalized approach is more robust to such bias compared with other systems, and have already started quantifying and addressing these issues as we find them in ongoing work.
In the meantime, we offer this paragraph as a disclaimer.

\section*{Acknowledgments}
We would like to thank Avi Shmidman for details about Dicta's Nakdan and other suggestions.
We thank Sara Gershuni for lengthy and fruitful discussions, and for her linguistic insights and advice.
We thank Yoav Goldberg, Reut Tsarfaty, Ian Stewart, Sarah Wiegreffe, Kyle Gorman and many anonymous reviewers for their comments and suggestions in discussions and on earlier drafts.

\bibliography{anthology,nakdimon}

\begin{thebibliography}{25}
\expandafter\ifx\csname natexlab\endcsname\relax\def\natexlab#1{#1}\fi

\bibitem[{Abandah et~al.(2015)Abandah, Graves, Al-Shagoor, Arabiyat, Jamour,
  and Al-Taee}]{abdanah-arabic}
Gheith Abandah, Alex Graves, Balkees Al-Shagoor, Alaa Arabiyat, Fuad Jamour,
  and Majid Al-Taee. 2015.
\newblock \href {https://doi.org/10.1007/s10032-015-0242-2} {Automatic
  diacritization of {A}rabic text using recurrent neural networks}.
\newblock \emph{International Journal on Document Analysis and Recognition
  (IJDAR)}, 18:183--197.

\bibitem[{Abu-Rabia(2001)}]{abu2001role}
Salim Abu-Rabia. 2001.
\newblock The role of vowels in reading semitic scripts: Data from {A}rabic and
  {H}ebrew.
\newblock \emph{Reading and Writing}, 14(1-2):39--59.

\bibitem[{Belinkov and Glass(2015)}]{belinkov-glass-2015-arabic}
Yonatan Belinkov and James Glass. 2015.
\newblock \href {https://doi.org/10.18653/v1/D15-1274} {{A}rabic diacritization
  with recurrent neural networks}.
\newblock In \emph{Proceedings of the 2015 Conference on Empirical Methods in
  Natural Language Processing}, pages 2281--2285, Lisbon, Portugal. Association
  for Computational Linguistics.

\bibitem[{Bentin and Frost(1987)}]{bentin1987processing}
Shlomo Bentin and Ram Frost. 1987.
\newblock Processing lexical ambiguity and visual word recognition in a deep
  orthography.
\newblock \emph{Memory \& Cognition}, 15(1):13--23.

\bibitem[{Blodgett et~al.(2020)Blodgett, Barocas, Daum{\'e}~III, and
  Wallach}]{blodgett-etal-2020-language}
Su~Lin Blodgett, Solon Barocas, Hal Daum{\'e}~III, and Hanna Wallach. 2020.
\newblock \href {https://doi.org/10.18653/v1/2020.acl-main.485} {Language
  (technology) is power: A critical survey of {``}bias{''} in {NLP}}.
\newblock In \emph{Proceedings of the 58th Annual Meeting of the Association
  for Computational Linguistics}, pages 5454--5476, Online. Association for
  Computational Linguistics.

\bibitem[{Gal(2002)}]{gal-2002-hmm}
Ya{'}akov Gal. 2002.
\newblock \href {https://doi.org/10.3115/1118637.1118641} {An {HMM} approach to
  vowel restoration in {A}rabic and {H}ebrew}.
\newblock In \emph{Proceedings of the {ACL}-02 Workshop on Computational
  Approaches to {S}emitic Languages}, Philadelphia, Pennsylvania, USA.
  Association for Computational Linguistics.

\bibitem[{Goldberg and Elhadad(2010)}]{goldberg-elhadad-2010-easy}
Yoav Goldberg and Michael Elhadad. 2010.
\newblock \href {https://www.aclweb.org/anthology/W10-1412} {Easy-first
  dependency parsing of modern {H}ebrew}.
\newblock In \emph{Proceedings of the {NAACL} {HLT} 2010 First Workshop on
  Statistical Parsing of Morphologically-Rich Languages}, pages 103--107, Los
  Angeles, CA, USA. Association for Computational Linguistics.

\bibitem[{Hochreiter and Schmidhuber(1997)}]{hochreiter1997long}
Sepp Hochreiter and J{\"u}rgen Schmidhuber. 1997.
\newblock Long short-term memory.
\newblock \emph{Neural computation}, 9(8):1735--1780.

\bibitem[{Kamir et~al.(2002)Kamir, Soreq, and
  Neeman}]{kamir-etal-2002-comprehensive}
Dror Kamir, Naama Soreq, and Yoni Neeman. 2002.
\newblock \href {https://doi.org/10.3115/1118637.1118646} {A comprehensive
  {NLP} system for modern standard {A}rabic and modern {H}ebrew}.
\newblock In \emph{Proceedings of the {ACL}-02 Workshop on Computational
  Approaches to {S}emitic Languages}, Philadelphia, Pennsylvania, USA.
  Association for Computational Linguistics.

\bibitem[{Kingma and Ba(2014)}]{kingma2014adam}
Diederik~P Kingma and Jimmy Ba. 2014.
\newblock Adam: A method for stochastic optimization.
\newblock \emph{arXiv preprint arXiv:1412.6980}.

\bibitem[{Kontorovich(2001)}]{kontorovichproblems}
Leonid Kontorovich. 2001.
\newblock Problems in {S}emitic {NLP}: {H}ebrew vocalization using {HMM}s.
\newblock In \emph{Problems in {S}emitic {NLP}, NIPS Workshop on Machine
  Learning Methods for Text and Images}.

\bibitem[{Moryossef et~al.(2019)Moryossef, Aharoni, and
  Goldberg}]{moryossef-etal-2019-filling}
Amit Moryossef, Roee Aharoni, and Yoav Goldberg. 2019.
\newblock \href {https://doi.org/10.18653/v1/W19-3807} {Filling gender {\&}
  number gaps in neural machine translation with black-box context injection}.
\newblock In \emph{Proceedings of the First Workshop on Gender Bias in Natural
  Language Processing}, pages 49--54, Florence, Italy. Association for
  Computational Linguistics.

\bibitem[{Mubarak et~al.(2019)Mubarak, Abdelali, Sajjad, Samih, and
  Darwish}]{mubarak-etal-2019-highly}
Hamdy Mubarak, Ahmed Abdelali, Hassan Sajjad, Younes Samih, and Kareem Darwish.
  2019.
\newblock \href {https://doi.org/10.18653/v1/N19-1248} {Highly effective
  {A}rabic diacritization using sequence to sequence modeling}.
\newblock In \emph{Proceedings of the 2019 Conference of the North {A}merican
  Chapter of the Association for Computational Linguistics: Human Language
  Technologies, Volume 1 (Long and Short Papers)}, pages 2390--2395,
  Minneapolis, Minnesota. Association for Computational Linguistics.

\bibitem[{N{\'a}plava et~al.(2021)N{\'a}plava, Straka, and
  Strakov{\'a}}]{naplava2021diacritics}
Jakub N{\'a}plava, Milan Straka, and Jana Strakov{\'a}. 2021.
\newblock Diacritics restoration using bert with analysis on czech language.
\newblock \emph{arXiv preprint arXiv:2105.11408}.

\bibitem[{N{\'a}plava et~al.(2018)N{\'a}plava, Straka, Stra{\v{n}}{\'a}k, and
  Haji{\v{c}}}]{naplava-etal-2018-diacritics}
Jakub N{\'a}plava, Milan Straka, Pavel Stra{\v{n}}{\'a}k, and Jan Haji{\v{c}}.
  2018.
\newblock \href {https://www.aclweb.org/anthology/L18-1247} {Diacritics
  restoration using neural networks}.
\newblock In \emph{Proceedings of the Eleventh International Conference on
  Language Resources and Evaluation ({LREC} 2018)}, Miyazaki, Japan. European
  Language Resources Association (ELRA).

\bibitem[{Plank et~al.(2016)Plank, S{\o}gaard, and
  Goldberg}]{plank-etal-2016-multilingual}
Barbara Plank, Anders S{\o}gaard, and Yoav Goldberg. 2016.
\newblock \href {https://doi.org/10.18653/v1/P16-2067} {Multilingual
  part-of-speech tagging with bidirectional long short-term memory models and
  auxiliary loss}.
\newblock In \emph{Proceedings of the 54th Annual Meeting of the Association
  for Computational Linguistics (Volume 2: Short Papers)}, pages 412--418,
  Berlin, Germany. Association for Computational Linguistics.

\bibitem[{Shacham and Wintner(2007)}]{shacham-wintner-2007-morphological}
Danny Shacham and Shuly Wintner. 2007.
\newblock \href {https://www.aclweb.org/anthology/D07-1046} {Morphological
  disambiguation of {H}ebrew: A case study in classifier combination}.
\newblock In \emph{Proceedings of the 2007 Joint Conference on Empirical
  Methods in Natural Language Processing and Computational Natural Language
  Learning ({EMNLP}-{C}o{NLL})}, pages 439--447, Prague, Czech Republic.
  Association for Computational Linguistics.

\bibitem[{Shmidman et~al.(2020)Shmidman, Shmidman, Koppel, and
  Goldberg}]{shmidman-etal-2020-nakdan}
Avi Shmidman, Shaltiel Shmidman, Moshe Koppel, and Yoav Goldberg. 2020.
\newblock \href {https://doi.org/10.18653/v1/2020.acl-demos.23} {{N}akdan:
  Professional {H}ebrew diacritizer}.
\newblock In \emph{Proceedings of the 58th Annual Meeting of the Association
  for Computational Linguistics: System Demonstrations}, pages 197--203,
  Online. Association for Computational Linguistics.

\bibitem[{Smith(2017)}]{smith2017cyclical}
Leslie~N Smith. 2017.
\newblock Cyclical learning rates for training neural networks.
\newblock In \emph{2017 IEEE Winter Conference on Applications of Computer
  Vision (WACV)}, pages 464--472. IEEE.

\bibitem[{Stankevi{\v c}ius et~al.(2022)Stankevi{\v c}ius, Luko{\v s}evi{\v
  c}ius, Kapo{\v c}i{\=u}t{\.e}-Dzikien{\.e}, Briedien{\.e}, and Krilavi{\v
  c}ius}]{Stankevicius2022}
Lukas Stankevi{\v c}ius, Mantas Luko{\v s}evi{\v c}ius, Jurgita Kapo{\v
  c}i{\=u}t{\.e}-Dzikien{\.e}, Monika Briedien{\.e}, and Tomas Krilavi{\v
  c}ius. 2022.
\newblock \href {http://arxiv.org/abs/2201.13242} {Correcting diacritics and
  typos with {ByT5} transformer model}.
\newblock \emph{arXiv e-prints}.

\bibitem[{Tomer(2012)}]{tomer2012automatic}
Eran Tomer. 2012.
\newblock \emph{Automatic Hebrew Text Vocalization}.
\newblock Ben-Gurion University of the Negev, Faculty of Natural Sciences,
  Department of Computer Science.

\bibitem[{Tsarfaty et~al.(2019)Tsarfaty, Sadde, Klein, and
  Seker}]{tsarfaty-etal-2019-whats}
Reut Tsarfaty, Shoval Sadde, Stav Klein, and Amit Seker. 2019.
\newblock \href {https://doi.org/10.18653/v1/D19-3044} {What{'}s wrong with
  {H}ebrew {NLP}? and how to make it right}.
\newblock In \emph{Proceedings of the 2019 Conference on Empirical Methods in
  Natural Language Processing and the 9th International Joint Conference on
  Natural Language Processing (EMNLP-IJCNLP): System Demonstrations}, pages
  259--264, Hong Kong, China. Association for Computational Linguistics.

\bibitem[{Vaswani et~al.(2017)Vaswani, Shazeer, Parmar, Uszkoreit, Jones,
  Gomez, Kaiser, and Polosukhin}]{vaswani2017attention}
Ashish Vaswani, Noam Shazeer, Niki Parmar, Jakob Uszkoreit, Llion Jones,
  Aidan~N Gomez, {\L}ukasz Kaiser, and Illia Polosukhin. 2017.
\newblock Attention is all you need.
\newblock In \emph{Advances in neural information processing systems}, pages
  5998--6008.

\bibitem[{Zalmout and Habash(2017)}]{zalmout-habash-2017-dont}
Nasser Zalmout and Nizar Habash. 2017.
\newblock \href {https://doi.org/10.18653/v1/D17-1073} {Don{'}t throw those
  morphological analyzers away just yet: Neural morphological disambiguation
  for {A}rabic}.
\newblock In \emph{Proceedings of the 2017 Conference on Empirical Methods in
  Natural Language Processing}, pages 704--713, Copenhagen, Denmark.
  Association for Computational Linguistics.

\bibitem[{Zhao et~al.(2017)Zhao, Wang, Yatskar, Ordonez, and
  Chang}]{zhao-etal-2017-men}
Jieyu Zhao, Tianlu Wang, Mark Yatskar, Vicente Ordonez, and Kai-Wei Chang.
  2017.
\newblock \href {https://doi.org/10.18653/v1/D17-1323} {Men also like shopping:
  Reducing gender bias amplification using corpus-level constraints}.
\newblock In \emph{Proceedings of the 2017 Conference on Empirical Methods in
  Natural Language Processing}, pages 2979--2989, Copenhagen, Denmark.
  Association for Computational Linguistics.

\end{thebibliography}
\clearpage

\appendix

\begin{table*}
    \centering
    \small
    \begin{tabular}{lrrrrrrrrrrrr} \toprule
         & \multicolumn{4}{c}{Dicta -- reported / reproduced} & & 
        \multicolumn{6}{c}{New test set (\S\ref{sec:test})} \\ & & & & & & & & & & \multicolumn{2}{c}{OOV} \\
        % \midrule
        System & \textsc{dec} & \textsc{cha} & \textsc{wor} & \textsc{voc} & & \textsc{dec} & \textsc{cha} & \textsc{wor} & \textsc{voc} & \textsc{wor} & \textsc{voc}\\ \midrule
        \multicolumn{2}{l}{~~~~Baselines} \\
        \textsc{MajMod} & 84.93 & 75.94 & 68.10 & 69.63 & & 88.04 & 81.22 & 76.14 & 77.10 & N/A & N/A \\
        \textsc{MajAll}    & 91.67 & 86.29 & 79.43 & 81.19 & & 93.79 & 90.01 & 84.87 & 86.19 & N/A & N/A\\
        \midrule
        \multicolumn{2}{l}{~~~~Lexicalized} \\
        \textsc{Snopi} & 87.81 & 78.96 / 79.92 & 66.41 / 66.57 & 70.35 & & 91.29 & 85.84 & 76.45 & 78.91 & 40.83 & 42.39 \\
        \textsc{Morfix} & 94.91 & 90.32 / 91.29 & 80.90 / 82.24 & 86.48 & & 96.84 & 94.92 & 90.38 & 92.39 & 63.91 & 69.20\\
        \textsc{Dicta} & \textbf{97.53} & \textbf{95.12 / 95.71} & \textbf{88.23 / 89.23} & \textbf{90.66} & & \textbf{97.95} & \textbf{96.77} & \textbf{94.11} & \textbf{94.92} & \textbf{76.21} & \textbf{77.66} \\
        \midrule
        \multicolumn{2}{l}{~~~~Unlexicalized} \\
        \sysname{}$_{0}$ & 95.78 & 92.59 & 79.00 & 83.01 & & 94.59 & 91.70 & 84.94 & 87.54 & 47.05 & 50.96 \\     % not updated
        \sysname{} & & & & & & 97.91 & 96.37 & 89.75 & 91.64 & 57.46 & 62.06 \\
        \bottomrule
    \end{tabular}
    \caption{Document-level macro \% accuracy on the test set from \newcite{shmidman-etal-2020-nakdan} and on our new test set. We cannot report our full \sysname{}'s performance on the former, as we use the test set for parts of its training.
    \textsc{MajAll} is reported as \textsc{Majority} in the main text; \textsc{MajMod} only considers text in the \textsc{Modern} portion of our training set.}
    % and on our new test set.}
    % \todo{Only report new test set, move rest to appendix}}
    % Results with $^\dagger$ are as reported in \newcite{shmidman-etal-2020-nakdan}.}
    % \todo{note about apples-oranges comparison, using nonstandard diacritization.}
    \label{tab:res}
\end{table*}

\section{Full Script Reconciliation}
\label{app:male}

We apply the following resolution tactics for added letters in undotted text:
(a) We almost never remove or add letters to the original text (unless it is completely undiacritizable).
(b) We keep \textit{dagesh} in letters that follow a \textit{shuruk} which replaces a \textit{kubuts}, and similarly for yod (\textit{hirik male} replacing \textit{hirik haser}).
(c) When we have double \textit{vav} or double \textit{yod}, the second letter is usually left undotted, except when it is impossible to have the correct vocalization this way.

Resolving \textit{ktiv haser} discrepancies from Morfix outputs is done by adding missing vowel letters, or removing superfluous vowel letters, in such a way that would not count as an error if it is correct according to Academy regulations.

\section{Development Experiments}
\label{app:dev}

We tried to further improve \sysname{} by initializing its parameters from a language model trained to predict masked characters in a large undotted Wikipedia corpus (440MB, 30\% mask rate), but were only able to achieve an improvement of 0.07\%. 
Attempted architectural modifications, including substituting a Transformer~\cite{vaswani2017attention} for the LSTM; adding a CRF layer to the decoding process; and adding a residual connection between the character LSTM layers, yielded no substantial benefits in these experiments.
Similarly, varying the number of LSTM layers between 2 and 5 (keeping the total number of parameters roughly constant, close to the 5,313,223 parameters of our final model) had little to no impact on the accuracy on the validation set.

\autoref{fig:part} shows the favorable effect of training \sysname{} over an increasing amount of \textsc{modern} text.
\begin{figure}
    \centering
    \includegraphics[width=0.8\columnwidth]{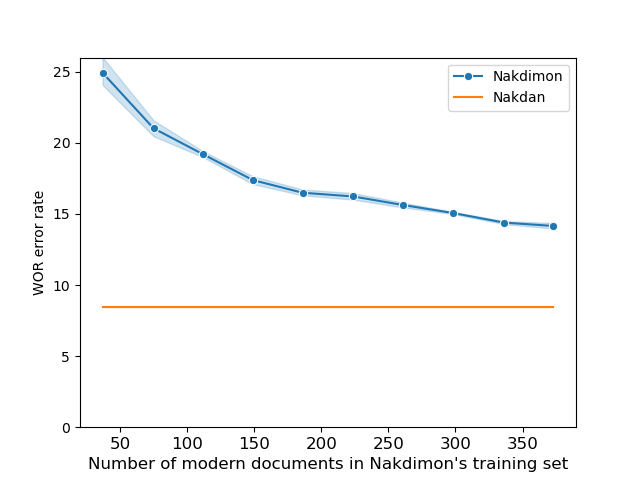}
    \caption{\textsc{wor} error rate on validation set as a function of training set size vs. Dicta, over five runs. Other metrics show similar trends.}
    \label{fig:part}
\end{figure}

\section{Dicta Test Set}
\label{app:results}

We present results for the Dicta test set in \autoref{tab:res}.
In order to provide fair comparison and to preempt overfitting on this test data, we ran this test in a preliminary setup on a variant of \sysname{} which was not tuned or otherwise unfairly trained.
This system, \sysname$_0$, differs from our final variant in three main aspects: it is not trained on the Dicta portion of our training corpus~(\S\ref{sec:train}), it is not trained on the \textsc{automatic} corpus, and it employs a residual connection between the two character Bi-LSTM layers. % results on dicta (full train leakage): 99.08  & 99.49 & 96.61 & 97.49
Testing on the Dicta test set required some minimal evaluation adaptations resulting from encoding constraints (for example, we do not distinguish between \emph{kamatz katan} and \emph{kamatz gadol}).
Thus, we copy the results reported in \newcite{shmidman-etal-2020-nakdan} as well as our replication.

We see that the untuned \sysname{}$_0$ performs on par with the proprietary Morfix, which uses word-level dictionary data, consistent with our main results on our novel test set.

\section{Hyperparameters}
\label{app:hyper}
We tuned hyperparameters and architecture over a held-out validation set of 40 documents with 27,681 tokens, on which Dicta performs at 91.56\% \textsc{wor} accuracy.

In our chosen setup, we train \sysname{} over \textsc{pre-modern} for a single epoch, followed by two epochs over the 
\textsc{automatic} corpus, and then by three epochs over the \textsc{modern} corpus.
We optimize using Adam~\cite{kingma2014adam}. For the \textsc{pre-modern} corpus we use a cyclical learning rate schedule~\cite{smith2017cyclical}, varying linearly from $3\cdot 10 ^{-3}$ through $8 \cdot 10^{-3}$ and down to $10^{-4}$, which we found to be more useful than a constant learning rate. For each of \textsc{automatic} and \textsc{modern} corpora we use epoch-wise decreasing learning rate: $(3\cdot 10^{-3}, 10^{-3})$ and $(10^{-3}, 10^{-3}, 3\cdot 10 ^{-4})$ respectively. We set maximum chunk size to 80 characters, and use batch size of 128.
% Based on tuning experiments, (\S\ref{ssec:dev}),
We set both character embedding and LSTM hidden dimensions to 400,
% (being cost efficient and round),
and apply a dropout rate of $0.1$.

\end{document}